\documentclass[11pt]{article} 
\usepackage{hyperref} 
\usepackage[margin=1.5in]{geometry}

\usepackage{url}
\usepackage{amsmath,amssymb} 
\usepackage{graphicx}
\usepackage{subfig}
\usepackage{filecontents}

\pdfoutput=1 
\begin{document} 
\title{Sub-threshold CMOS Spiking Neuron Circuit Design \\ for Navigation  \\ Inspired by \textit{C. elegans} Chemotaxis } 
\author{Shibani Santurkar and Bipin Rajendran \\ 
 \\\vspace{6pt} \hspace{-0.5in} Deparment of Electrical Engineering, 
Indian Institute of Technology Bombay, India \\
shibani@iitb.ac.in, bipin@ee.iitb.ac.in} 
\maketitle 
\begin{abstract} 
We demonstrate a spiking neural network for navigation motivated by the chemotaxis network of \textit{Caenorhabditis elegans}. Our network uses information regarding temporal gradients in the tracking variable's concentration to make navigational decisions. The gradient information is determined by mimicking the underlying mechanisms of the ASE neurons of  \textit{C. elegans}. Simulations show that our model is able to forage and track a target set-point in extremely noisy environments. We develop a VLSI implementation for the main gradient detector neurons, which could be integrated with standard comparator circuitry to develop a robust circuit for navigation and contour tracking.
\end{abstract} 

\section{Introduction}
\label{sec:intro}
Spiking Neural Networks (SNNs) form a special sub-class of Artificial Neural Networks (ANNs) which use the  timing of arrival of spikes as the main mechanism for communication and computation. It has been shown that SNNs have higher computational capabilities as compared to previous generations of NNs for performing many computational tasks \cite{Maass19971659}. SNNs allow for rapid decoding of information as they rely on the timing of individual spikes rather than average firing rate. Further, hardware implementations of large SNNs are easier as all computations are performed based on binary spikes in an event-triggered manner. There is hence significant interest to develop SNN based circuits to  solve various engineering problems such as signal processing, pattern recognition \& classification  and navigation control.

All higher biological organisms employ neurons that issue spikes for cognitive tasks. However, the enormity of the network size of these systems has  motivated the use of organisms with simpler connectivity statistics such as \textit{Caenorhabditis elegans} \cite{Brenner01051974} for experimental analysis to decipher the computational motifs of nature. \textit{C. elegans}, with $302$ neurons and $\approx$ $5000$ chemical synapses, $2000$ neuromuscular junctions and $600$ gap junctions \cite{White12111986}, has presumably one of the simplest and most well-understood nervous systems today. Despite this simplicity, it shows sophisticated functionality with the ability to perform chemotaxis \cite{Ward01031973}-\cite{Dusenbery01051973}, thermotaxis, etc  and to learn, adapt and remember.

In this paper we develop a SNN which can track contours of physical variables such as chemical concentration, temperature, etc. inspired by the NaCl chemotaxis circuit in  \textit{C. elegans}. We create an artificial model of a spiking neuron pair inspired by the underlying dynamics of ASE neurons in \textit{C. elegans}, which is believed to play a pivotal role in the chemotaxis to NaCl. We have created a  simple network model incorporating these neurons, that can effectively perform contour tracking. We study the performance of our network for various concentration profiles as well as in noisy environments. We then develop a VLSI circuit design for the ASE neurons based on analog sub-threshold CMOS circuits, which could be integrated with other standard neuronal circuitry for achieving energy efficient hardware for performing navigation and contour tracking.

\section{Modeling ASEL and ASER dynamics}
\label{sec:model}

Laser ablation experiments  show that one specific neuron pair denoted as the ASE neurons, when ablated cause \textit{C. Elegans} to have severely reduced chemotaxis towards NaCl \cite{Bargmann1991729}. It is widely believed that the ASE neuron pair is crucial towards chemotaxis with residual functionality spread over numerous other neurons. There is experimental evidence to support that the ASEL neuron responds to up-steps and the ASER responding to down-steps in NaCl concentration  \cite{Bargmann1991729}- \cite{Miller30032005}. Most published literature on neural networks inspired by chemotaxis of the \textit{C. elegans} doesn\rq{}t accommodate this key neuron pair and most capture information regarding the local NaCl concentration in a current input to a single sensory neuron. The model we have implemented, which is based on \cite{PA}, captures information about the local  concentration of the tracking variable is captured via depolarizing and hyperpolarizing ion channels. The membrane potential of ASE neurons is modeled as
\begin{equation} \label{eq:pot}
\tau_m \dot{V} = (V_0 - V) +  k^d (V_d - V) + k^h (V_h - V)
\end{equation}
$\tau_m$ is the membrane time constant, $V_0$ is the resting membrane potential, $k^{d,h}$ and $V_{d,h}$ capture the conductivity and reversal potential of the depolarizing and hyperpolarizing ion channels respectively.

The depolarizing ion channels follow a three-state model - an initial unbound state, the conducting bound state and the final inactive state. Transitions from the unbound to bound state are triggered when the local concentration exceeds/goes below the threshold concentration $C_{L/R}$ of the ASEL/R neuron. The thresholds, $C_{L,R}$, adapt to the tracking variable\rq{}s local concentration, without which the worm would not  chemotax. As the fraction of channels in the bound state increases, conductivity of depolarizing channels and membrane potential of the neuron increases. The magnitude of the response of ASEL/R to a step in concentration depends on the transition rate from unbound to bound state which is proportional to the concentration step. The hyperpolarizing ion channels, which are present only in the ASER, follow a simpler two-state model - an initial unbound state and a conducting bound state. When the local concentration is greater than the threshold $C_R$, these ion channels transition to the bound state and the conductivity of the hyperpolarizing channels increases causing a decrease in the membrane potential. These channels do not adapt to the local concentration of the tracking variable, i.e, $\alpha^h$ is not a function of local concentration (C) unlike $\alpha^d$. 
The state transition equations for the depolarizing and hyperpolarizing ion channels respectively are 
\begin{align}
\label{eq:dep}
\left[ \begin{matrix}
\dot{u^d} \\
\dot{b^d} \\
\dot{i^d}
\end{matrix}\right]
&= \left[ \begin{matrix}
- \alpha^d & \beta^d & \delta^d\\
\alpha^d & - \beta^d - \gamma^d & 0\\
0 &  \gamma^d & -\delta^d \end{matrix}\right]
\left[ \begin{matrix}
u^d \\
b^d  \\
i^d 
\end{matrix}\right] \ \ \ \ \ \ \ \ \ 
\left[ \begin{matrix}
\dot{u^h}  \\
\dot{b^h} 
\end{matrix}\right]
= \left[ \begin{matrix}
- \alpha^h & \beta^h\\
\alpha^h & - \beta^h \end{matrix}\right]
\left[ \begin{matrix}
u^h \\
b^h  
\end{matrix}\right]
 \end{align}
The dimensionless variable, $k^{d,h}$, that captures conductivity dependence  is modeled as 
\begin{equation}
\label{eq:cond}
k^{d,h} = k_m \times ( b^{d,h} )^2 
\end{equation}
The transition rates $\alpha^h,\beta^{d,h}, \gamma^{d}, \delta^{d}$ are constants. 

The rate $\alpha^d$ determines the response of ASEL/R to an up/down-step and is modeled as 
\noindent\begin{minipage}{ \linewidth}
\begin{align}
\label{eq:alphad}
\alpha_L^d &= \alpha_{L0}^d (C - C_L) \times H(C - C_L) &  
\alpha_R^d &= \alpha_{R0}^d (C - C_R) \times H(C_R - C)
\end{align}
\end{minipage}   
where $\alpha_{L0}^d $ and $\alpha_{R0}^d $ are scaling factors. The threshold concentrations-$C_{L,R}$ adapt to the ambient concentration C according to the adaptation rules given by:
\begin{equation} \label{eq:asel_thresh}
\begin{array}{l}
\dot{C_{L}} = (C \times H(C - C_L) - C_{L})/\tau_{L} \\
\dot{C_{R}} = (C \times H(C_R - C) - sgn(C_R - C) C_{R})/\tau_{R}  
\end{array}
\end{equation}
For the ASER neuron we have modified the adaptation model from that presented in \cite{PA}. To ensure that threshold $C_R$ does\rq{}t get stuck at 0, we must impose a minimum on the threshold value, i.e.,
\begin{equation}
C_R = max(C_R, C_{R,min})
\end{equation}
where the value of $C_{R}$ is given by  \ref{eq:asel_thresh}
The parameter $\alpha^h$ is modeled as
$\alpha_{R}^h = \alpha_{0}^h H(C - \eta_{R})$
where $\eta_R$ is the threshold concentration above which hyperpolarizing ion channels are active for the ASER neuron and $H(x)$ is the Heaviside function and $sgn(x)$ is the Signum function. 
\begin{figure}[!t]
\begin{center}
    \subfloat[
\label{fig:biodata}]{%
      \includegraphics[width=0.35\textwidth]{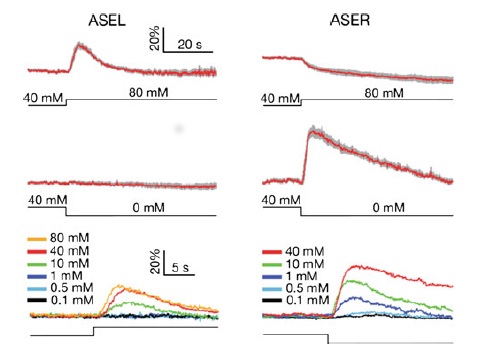}
    }
    \hspace{5em}
    \subfloat[
\label{fig:aselr_resp}]{%
      \includegraphics[width=0.27\textwidth]{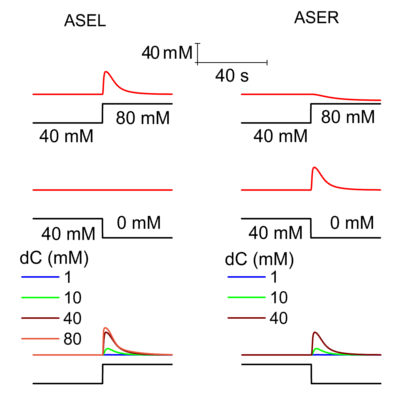}
    }
\end{center}
 \caption{
(a) Calcium imaging data showing the response of ASE neurons when subjected to different step profiles in concentration of NaCl (Figure adapted from \cite{Suz})
(b) Response of numerically simulated ASE neurons. \textit{Top}: Response of ASE neuron pair to an up-step in concentration.
\textit{Middle}: Response of ASE neuron pair to a down-step in concentration.
\textit{Bottom}: Response of ASE pair to steps of different magnitudes from a baseline concentration of $40\,$mM.}
  \end{figure}

Figure \ref{fig:biodata} shows results of  experiments to test the response of the ASE neurons to different concentration up-steps and down-steps. Figure \ref{fig:aselr_resp} shows the response of the numerically simulated neurons to  similar concentration profiles. As can be seen, there is excellent agreement between the model and experimentally observed behavior. Figure \ref{fig:aselr_resp}    shows the behavior of the neurons when presented with different concentration gradients. A strong change in membrane potential is observed for sharper and stronger gradients in input concentration.

\section{Modeling the Chemotaxis Network}
 
In standard chemotaxis,  \textit{C. elegans} navigates towards it\rq{}s cultivation concentration, or towards a concentration where it received food in the past \cite{kun}.
In order to convert the sensory cues into directed movement, it is believed that \textit{C. elegans} adopts two key strategies -  (i) {\textit{Klinokinesis}} or {\textit{biased random walk}}\cite{Pierce-Shimomura01111999}-\cite{Pierce-Shimomura15122005} and (ii)
{\textit{Klinotaxis}} which is the sinusoidal movement of the worm biased towards higher attractant concentration.
The worm shows no bias in klinokinesis when it is placed at concentration equal to the desired set-point, but it shows weakly negative klinotaxis \cite{kun}.

\begin{figure}[t!]
    \subfloat[
\label{fig:circuit}]{%
      \includegraphics[width=0.36\textwidth]{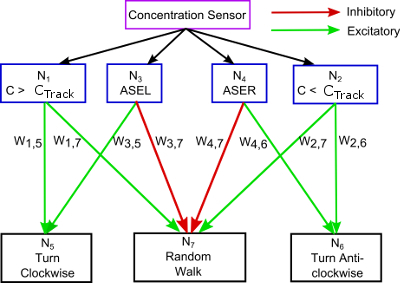}
    }
    \subfloat[
\label{fig:ase}]{%
      \includegraphics[width=0.32\textwidth]{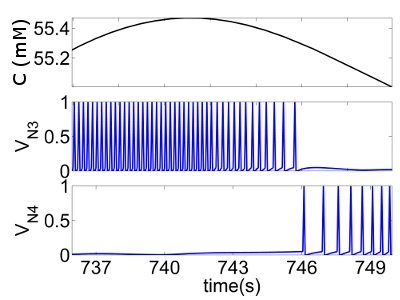}
    }
    \subfloat[
\label{fig:spike_freq}]{%
      \includegraphics[width=0.33\textwidth]{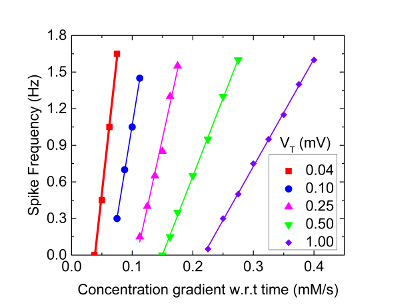}
    }
 \caption{
(a) Block Diagram of bio-inspired contour tracking network which takes input from a single concentration sensor to steer the worm towards desired set-point $C_{Track}$.
(b) Normalised response of gradient detector neurons - $N_3$ and $N_4$, inspired by ASE neurons, when subjected to a certain concentration profile.  $N_3$ and $N_4$ act as positive and negative gradient detectors respectively.  
(c) Spike frequency of the gradient detector neurons as a function of temporal gradient in concentration. As $V_T$ decreases, neurons become more sensitive to lower gradients in concentration.
}
  \end{figure}

As mentioned in Section \ref{sec:intro}, SNNs offer many advantages over their non-spiking counterparts motivating the transformation the ASE neuron pair developed in Section \ref{sec:model} into spiking neurons. These neurons can then be used in conjunction with other spiking neuron models, for instance the leaky-integrate-and-fire (LEIF) neurons to develop a SNN model for contour tracking. 
We now discuss the essential aspects of the navigation control for our \lq\lq{}worm\rq\rq{}, which relies on information about temporal gradients in concentration sensed (C) to make decisions in order to track a desired target set-point ($C_{Track}$).\\
$\bullet$ When the  \lq\lq{}worm\rq\rq{} is on a roughly flat surface (in terms of concentration) away from $C_{Track}$, it should rapidly explore the space by  random walk or foraging until a favorable direction is identified. \\
$\bullet$ When moving in an unfavorable direction, away from $C_{Track}$, it should alter its direction of motion, similar to the rapid pirouettes in \textit{C. elegans}. In our design, we chose to assign a clockwise turn when  moving up the gradient ($dC/dt > 0$) and it is already above the set-point ($ C > C_{Track}$) and an anti-clockwise turn in the opposite case when $dC/dt < 0$ and $ C < C_{Track}$ . \\
$\bullet$ When the worm is moving in a favorable direction towards $C_{Track}$, i.e., (i) $dC/dt > 0$ and $ C < C_{Track}$, (ii) $dC/dt < 0$ and $ C > C_{Track}$ or (iii) $C = C_{Track}$, the direction of motion should be unaltered, similar to the rare pirouettes in  \textit{C. elegans}. The overall block diagram for our network is shown in Figure \ref{fig:circuit}.

\subsection{Concentration Sensing Neurons}
\label{subsec:conc}

We employ two concentration sensing neurons, $N_1$ to detect if concentration is above the set-point ($C > C_{Track}$), and $N_2$ to detect if concentration is below the set-point ($C <  C_{Track}$). These neurons are modeled as LEIF neurons, where  the membrane potential $V(t)$ is governed by 
\begin{equation}
\begin{array}{l}
C \dot{V(t)} = -g_L(V(t) -  V_0) + I_{app}(t) + I_{syn}(t); \ \ \ 
If \  V(t)\geq V_T, V(t) \rightarrow V_{max}, V(t+dt) \rightarrow V_0
\end{array}
\end{equation}
C and $g_L$ are the membrane capacitance and conductance respectively, $V_0$ is the resting potential, $V_T$ is the threshold voltage. $I_{app}(t)$ represents the externally applied current and $I_{syn}(t)$ represents current due to synaptic connections with other neurons. $I_{syn}(t)$ due to a spike at time $t^k$ is given by 
\begin{equation}
I_{syn}(t) = I_0 \times w_{synapse}\times [e^{-(t-t^k)/\tau} - e^{-(t-t^k)/\tau_s}]
\end{equation} where $w_{synapse}$ denotes the strength (weight) of the synapse, and $\tau$ and $\tau_s$ are characteristic time constants of the synapse. The concentration sensing neurons are modeled as independent input neurons receiving only external input current and zero synaptic current from any other neurons in the network. The input current for $N_1$ and $N_2$ respectively is modeled as
\begin{equation}
\begin{array}{l}
I_{app,1}(t) = I_{app,0}  H(C - C_{Track}) \ \ \ \ \ \ \
I_{app,2}(t) = I_{app,0}  H(C_{Track} -  C)
\end{array}
\end{equation}
$N_1$ and $N_2$ spike at a fixed frequency if concentration is greater (or lesser) than the threshold. The simplicity of these two input neurons makes it extremely easy to tune the set-point and hence our worm is able to track any specified concentration. Note that these input neurons could be sensing concentration or intensity of any physical variable such as noxious gases, radiation, temperature etc.

\subsection{Gradient Detectors}
We develop spiking gradient detector neurons $N_3$ and $N_4$, whose spike-frequency encodes information about the temporal gradient of the concentration using the underlying dynamics of the  ASE neurons of \textit{C. elegans}. While computing $V(t)$ using equation \ref{eq:pot}, we apply this additional constraint
\begin{equation}
If \ \ V(t) \geq V_T, V(t) = V_{max}, V(t+dt) = V_0
\end{equation}
Figure \ref{fig:ase} shows the response of $N_3$ and $N_4$ to an arbitrary concentration profile and captures the dependence of spike frequency on the temporal gradient. The variation of spike frequency for the ASE neurons as a function of temporal derivative of concentration is shown for different $V_T$ values in Figure~\ref{fig:spike_freq}. We observe that as $V_T$ decreases, smaller temporal gradients can be sensed, making $N_3$ and $N_4$ very versatile gradient detectors. This provides a simple mechanism to modulate the sensitivity and performance of our model by simply controlling $V_T$.

\subsection{Navigation Control}
$\bullet$ The \lq\lq{}worm\rq\rq{} needs to sense when it is moving in the \lq\lq{}wrong\rq\rq{} direction, i.e., away from $C_{Track}$, so that it can alter its direction, for which we use two LEIF neurons $N_5$ and $N_6$. 
$N_5$ receives a constant negative bias current $I_{bias,5}$ and is connected to neurons $N_1$ and $N_3$ via excitatory synapses. The bias current is applied such that $N_5$ spikes if and only if both $N_1$ and $N_3$ spike. This ensures that $N_5$ spikes only if the positive gradient detector is spiking and local concentration is greater than $C_{Track}$. The worm then makes a turn in the clockwise direction with an angle which we chose to be $3.33^\circ$. $N_6$ is similarly designed with a negative bias current $I_{bias,6}$ and excitatory connections from $N_2$ and $N_4$ such that it would spike if the concentration sensed is less than the set-point and the worm is moving down the gradient. In this case, the worm makes an anti-clockwise turn with an angle of $3.33^\circ$. The spiking of $N_5$ and $N_6$, and hence the turning of the worm depends on the spike rate of $N_3$ and $N_4$ which in turn depends on the temporal gradient. Therefore if the worm deviates a large amount from $C_{track}$, there will be a stronger tendency to turn and return to a favorable course. \\
$\bullet$ The \lq\lq{}worm\rq\rq{} should make rapid exploratory motion when it is \lq\lq{}lost\rq\rq{}. 
It can be considered \lq\lq{}lost\rq\rq{} when not at $C_{Track}$ and it\rq{}s gradient detectors are not spiking, i.e., when not receiving any feedback whether it is on a favorable or unfavorable course. In our model, the worm makes decisions to random walk based on the spiking of $N_7$ which is implemented as an LEIF neuron.
$N_7$ has excitatory synapses from $N_1$ and $N_2$ and inhibitory synapses from $N_3$ and $N_4$. $N_1$ and $N_2$ spiking indicates the concentration sensed does not match the set-point. $N_3$ and $N_4$ not spiking indicates that the worm is on a flat concentration profile, with gradient less than detection threshold of the gradient detectors and hence must randomly explore to find a favorable direction. When $N_7$ spikes, the worm makes a turn with a  randomly chosen angle from the interval $\pm22.5^\circ$. \\
$\bullet$ While the worm is in a phase of random exploration, i.e., when $N_7$ is spiking, it is desirable that the worm rapidly explore a large area and hence the velocity of the worm is chosen to be relatively high, $v_1 = 0.3\,$mm/s. When $N_5$ or $N_6$ spike, the worm moves with a reduced velocity of $v_2 = 0.09\,$mm/s so as to reduce deviation about $C_{Track}$ and  improve tracking accuracy.  The velocity of the worm is thus either $v_1$ or $v_2$ depending on which neuron spiked last.
\section{Results}
\subsection{Simulation Results}
\begin{figure}[t!]
    \subfloat[
\label{fig:forage_55}]{%
      \includegraphics[width=0.34\textwidth]{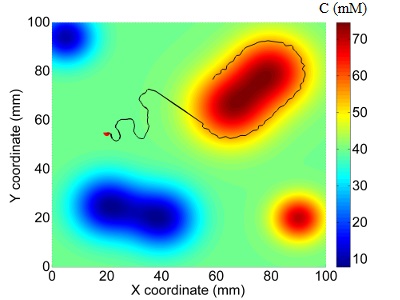}
    }
    \subfloat[
\label{fig:ons}]{%
      \includegraphics[width=0.34\textwidth]{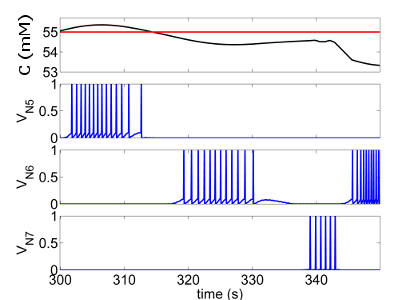}
    }
    \subfloat[
\label{fig:noise_ct}]{%
      \includegraphics[width=0.34\textwidth]{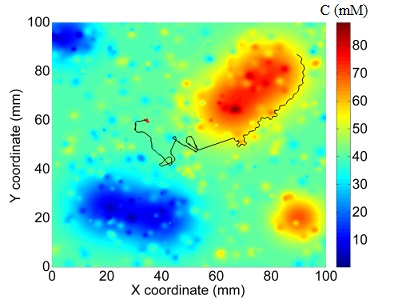}
    }
 \caption{
(a) Trajectory of our worm from initial position (red dot) to $C_{track} = 55\,$mM (b) Spike patterns of $N_5$, $N_6$ and $N_7$ while tracking desired set-point (c) Performance of the worm in a noisy environment, illustrating the robustness of our algorithm in the presence of salt and pepper noise.
}
  \end{figure}

Figure~\ref{fig:forage_55} shows a typical track of motion traced by our model worm. The arena for exploration is a $10\,$cm $\times 10\,$ cm  square plate with several hills and valleys of  concentration ranging from $10\,$mM to $70\,$mM. The worm starts its track from a roughly flat region with concentration $40\,$mM, with desired set-point for tracking $C_{Track}= 55\,$mM. It initially performs random exploration in pursuit of a favorable direction. Once a favorable direction is detected, it travels straight till it reaches the vicinity of the desired set-point. Subsequently it tracks the desired set-point with an accuracy of about $\approx 0.6\,$mM.  Figure \ref{fig:ons} shows the response of the navigation control neurons, $N_5$, $N_6$ and $N_7$ to an exemplary concentration profile during this track of the worm. $N_5$ and $N_6$ spike when the worm is moving  away from set-point and $C > C_{Track}$ or $C < C_{Track}$ respectively. 
$N_7$ spikes when the worm is on an almost flat concentration profile away from the set-point and causes the worm to random walk. It has been shown previously \cite{Pappleby2013} that the performance degrades drastically in the presence of noise, if non-spiking ASE neurons \cite{PA} are used to build similar tracking networks. Figure ~\ref{fig:noise_ct}  shows the performance of our model in an extremely noisy environment, with absolute value of noise in the range of $\approx 0-12\,$mM. Despite the environment being noisy, the worm is able to track the contour effortlessly. 
\subsection{Performance Evaluation}

To evaluate the performance of our worm to identify a set-point by foraging, we perform several experiments where the worm starts from the same initial position with the tracking set-point set to $55\,$mM. For the 200 simulations performed, the model worm identified the tracking set-point in  $89.5\%$ cases within $1500\,$s. Further, we observed that in $63\%$ of the cases, our worm reaches the desired set-point in under $530\,$s. Once the worm locates the desired set-point, it tracks it with an average deviation of $0.6\,$mM. In order to evaluate the efficiency of our foraging strategy, we compare it to the optimal search strategy for finding randomly distributed targets. This entails making flight-lengths between random turns follow the heavy-tailed L\'{e}vy distribution \cite{nolan:2015}. We simulated this strategy by drawing run lengths from a truncated L\'{e}vy distribution with $P(l) \propto l^{-2}$ in the interval $[s_{min}, s_{max}]$. The $s_{min}$ and $s_{max}$ were determined empirically from the neuron model, with $s_{min}=0.2649\,$mm being the most probable run-length for the neuron model and $s_{max} = 40\,$mm being the maximum flight-length for the neuron model. The set-point was reached in only $21\%$ of the cases within $1500\,$s as opposed to  $89.5\%$ in the case of our model. The success criteria for foraging was set as the track reaching within $0.5\,$mM of the set-point.

\section{VLSI Circuit Design for ASEL Neuron}
In order to show that the complex set of differential equations necessary to model the dynamics of the ASE neurons can be implemented in energy efficient hardware, we 
designed a circuit block  to capture the non-spiking graded potential response described in section \ref{sec:model}. The basic processing elements in the circuit implementation are $180\,$nm CMOS transistors biased in the sub-threshold regime.  We use variants of the basic Tau cell \cite{schaik2003} to implement differential equations in our circuit, with all the variables represented as currents.  In these circuits, the current relationship, as dictated by the translinear principle is
\begin{equation}
\label{eq:cr}
I_{out} \times (C \dot{V_C} + I_r) = I_{in} \times I_r \\ 
\end{equation}
where $V_C$ is the capacitor voltage. If $ I_r  \triangleq   CV_t/(k \tau)$, this circuit can be exploited to implement
\begin{equation}
\label{eq:de}
\tau \dot{I_{out}} = I_{in} - I_{out}
\end{equation}
The major sub-blocks of our circuit implementation are : \\
$\bullet$ \textit{\textbf{Adaptation Block:}} \\
Equation \ref{eq:asel_thresh} captures the adaptation of the threshold concentration $C_L$ to the ambient tracking variable  concentration, which is vital for chemotaxis. To model this behaviour in our circuit we compare Equation \ref{eq:asel_thresh} with \ref{eq:de} and observe that if $I_{out}$ represents $C_L$, then 
\begin{equation}
\label{eq:iin}
I_{in} = C \times H(C - I_{out})
\end{equation}


\begin{figure}[!t]
\begin{center}
\includegraphics[scale=0.35]{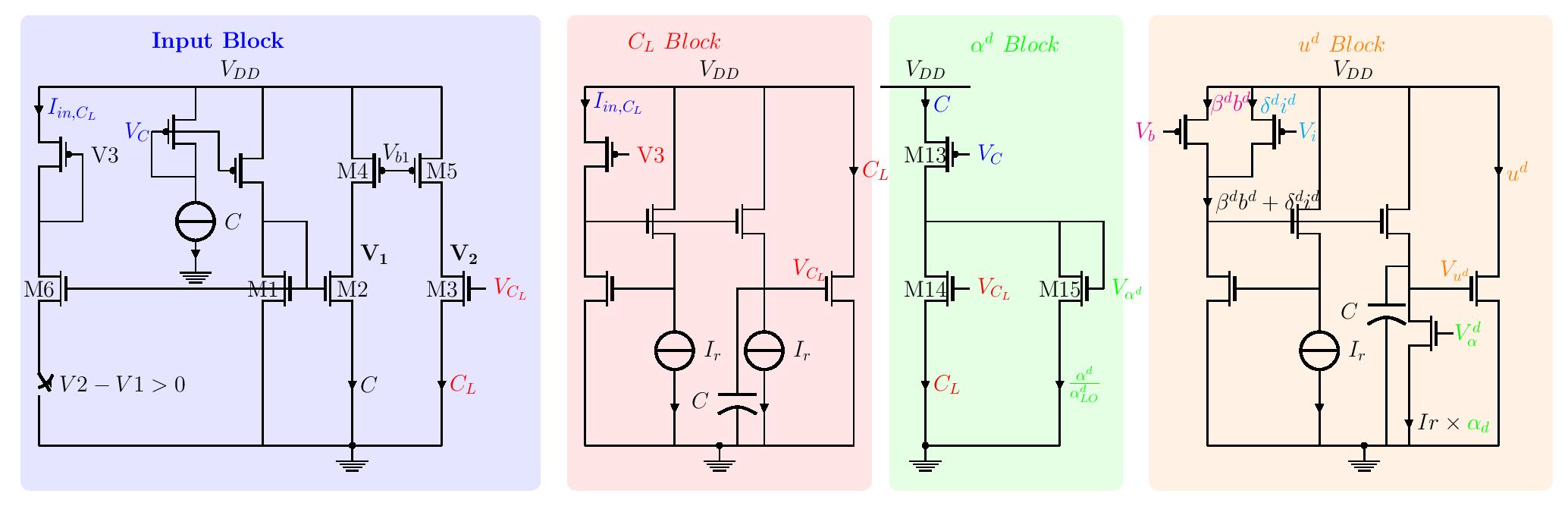}
\end{center}
 \caption{ Sub-circuit of VLSI implementation for ASEL neuron comprising the Input Block, $C_L Block$, $\alpha^d Block$ and $u^d Block$. Connections between blocks are not shown explicitly but through appropriate labels at gates of connected transistors.}
\label{fig:c1} 
\end{figure}
\begin{figure}[!t]
\begin{center}
\includegraphics[scale=0.33]{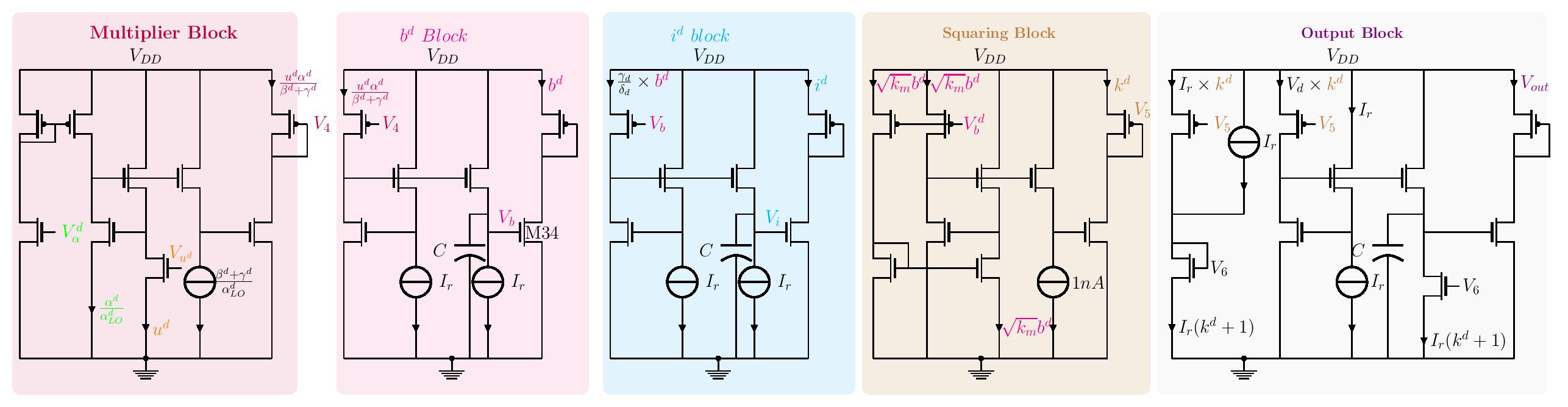}
\end{center}
\caption{Sub-circuit of VLSI implementation for ASEL neuron comprising the Multiplier Block, $b^d Block$, $i^d Block$, Squaring Block and Output Block. Connections between blocks are not shown explicitly but through appropriate labels at gates of connected transistors.}
\label{fig:c2}
\end{figure}


where C represents the input current coming from the concentration sensor. This input current is generated by the $Input Block$ in Figure \ref{fig:c1} wherein M1 carries current coming from concentration sensor with information about the tracking variable, M2 carries a mirrored version of this current and M3 carries a mirrored version of $I_{out} (\equiv C_L)$. M4 and M5 act as resistors (R) and the voltage difference $V_2 - V_1$, which represents $ (C - I_{out}) R$ is used to control an ideal switch. When the switch is turned on ($V_2 - V_1 > 0$), current through M6 is C since M6 mirrors current from M1. When the switch is off, the current through M6 is zero. Hence the current flowing through M6 represents $I_{in}$ as governed by Equation \ref{eq:iin}, represented in the figures as $I_{in,C_L}$. The $C_L Block$ of Figure \ref{fig:c1} captures the adaptation characteristics of Equation \ref{eq:asel_thresh} to evaluate $C_L$. \\
$\bullet$ \textit{\textbf{Computing $\mathbf{\alpha^d}$:}} \\
The parameter $\alpha^d$ is crucial in the rate transition equations since it determines the graded response of the ASEL neuron due to its dependence on the concentration sensed by the worm as shown in Equation \ref{eq:alphad}. In the $\alpha^d Block$ of Figure \ref{fig:c1} current through M13 is obtained by mirroring the input current C and current through M14 is a mirrored version of $C_L$, which is the output of $C_L Block$. As a result, current through the NMOS transistor M15 is $C - C_L$ if $C > C_L$ and 0 otherwise as required and hence represents ${\alpha^d}/{\alpha_{LO}^d}$, where ${\alpha_{LO}^d}$ is a constant scaling factor. \\
$\bullet$\textit{\textbf{Rate Transition Equations:}}\\
We now develop a circuit to implement the rate transition equations that govern the behavior of the depolarizing ion channels (Equation \ref{eq:dep}). The differential equation for unbound state, $u^d$ is given by
\begin{equation}
\label{eq:ud}
\dot{u^d} = (\beta^d b^d + \delta^d i^d) - (\alpha^d u^d)
\end{equation}
Comparing Equations \ref{eq:ud} and \ref{eq:de}, the translinear relationship needed to implement Equation \ref{eq:ud} is
\begin{equation}
u^d \times (C \dot{V_x} + \alpha^d I_r) = (\beta^d b^d + \delta^d i^d) \times I_r \\ 
\end{equation}
with $u^d$ as the output current of this circuit and $\beta^d b^d + \delta^d i^d$ is the input current. We have chosen to accelerate the performance of our circuit by a factor of $\sim 10^6$, and hence  the characteristic time scale of our circuit is in the order of a few micro-seconds. This is done by changing the scale of all time dependent parameters such as time constants, rate constants and velocity to micro-seconds instead of seconds. Hence in order to implement Equation \ref{eq:ud}, $\tau = 1\,\mu$s and the values of C and $I_r$ are taken to obtain this $\tau$ while keeping the transistors in the sub-threshold regime. The $u^d Block$ of Figure \ref{fig:c1} implements Equation \ref{eq:ud}. 
In order to determine $b^d$, the relevant differential equation is
\begin{equation}
\label{eq:bd}
1/(\beta^d + \gamma^d) \dot{b^d} = (\alpha^d u^d)/(\beta^d + \gamma^d) - b^d
\end{equation}
which on comparison with Equation \ref{eq:de} gives $I_{in} = (\alpha^d u^d)/(\beta^d + \gamma^d)$ and  $\tau = 1/(\beta^d + \gamma^d)$. The product $(\alpha^d u^d)$ is computed by the circuit  shown in $Multiplier Block$ of Figure \ref{fig:c2}. 
The $b^d Block$ of Figure \ref{fig:c2} implements the above differential equation. A similar methodology is used to design the block to compute $i^d$ which is shown in Figure \ref{fig:c2}


\begin{figure}[!t]
    \subfloat[
\label{fig:nr}]{%
      \includegraphics[width=0.34\textwidth]{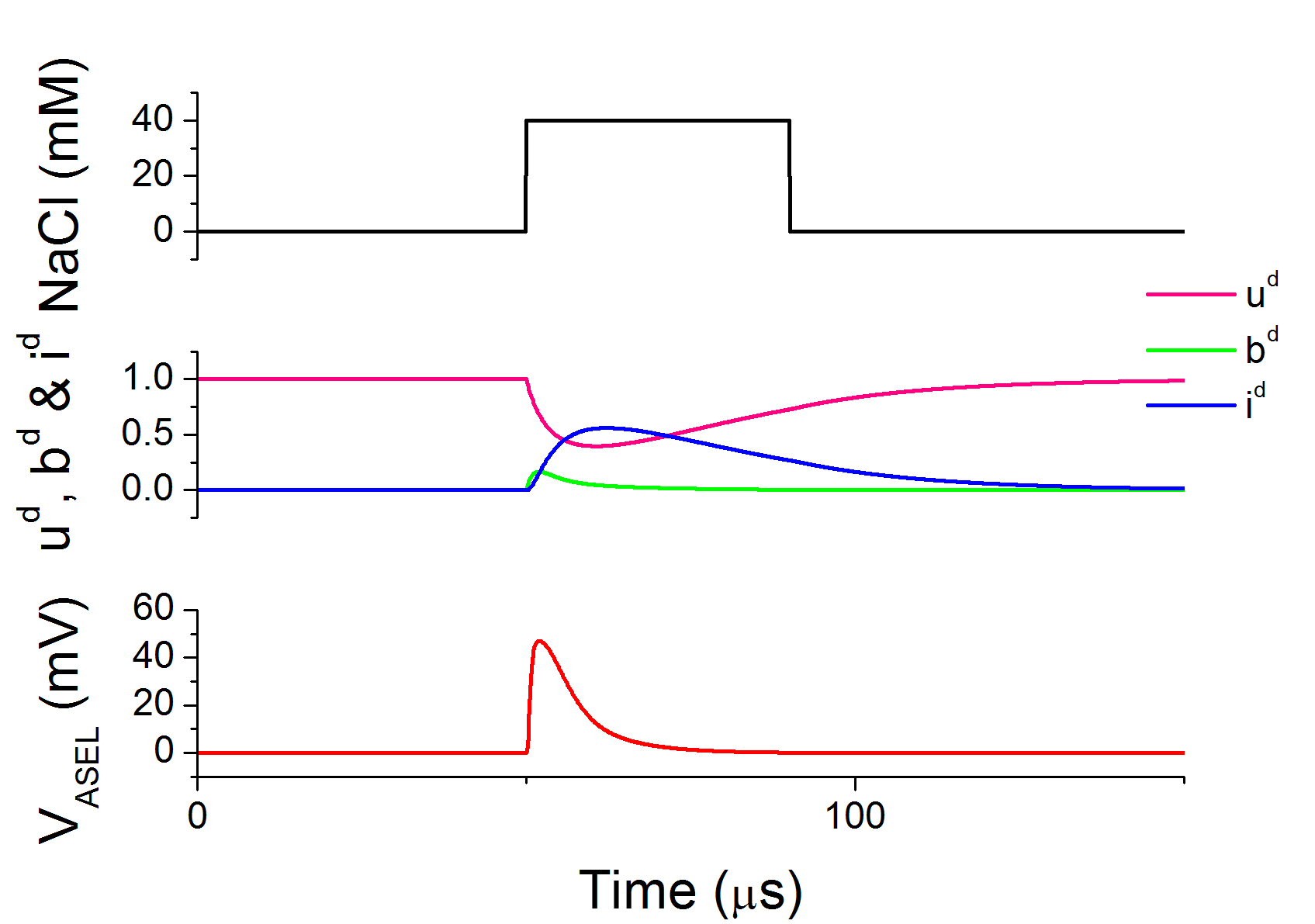}
    }
    \subfloat[
\label{fig:cr}]{%
      \includegraphics[width=0.34\textwidth]{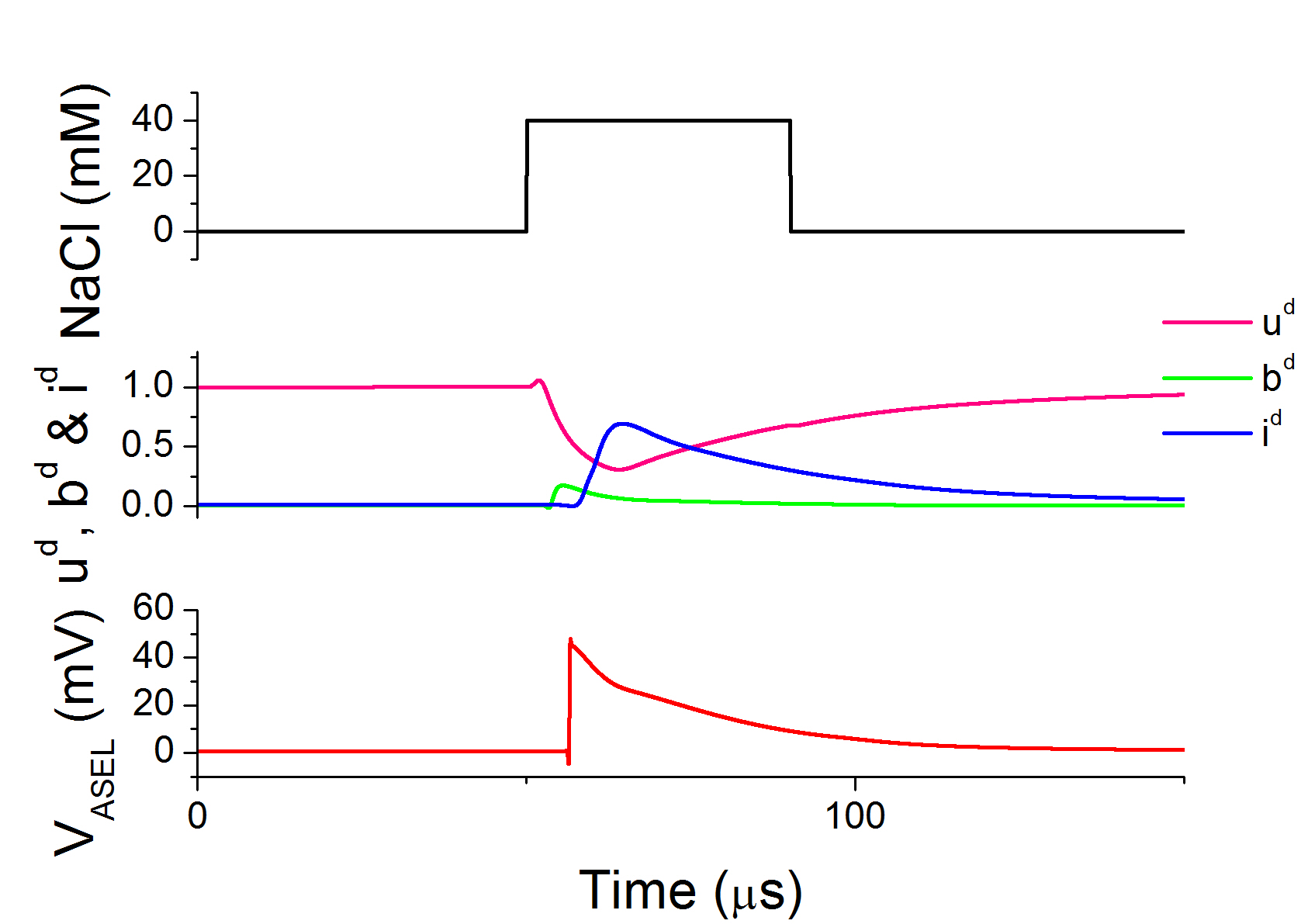}
    }
    \subfloat[
\label{fig:steps}]{%
      \includegraphics[width=0.33\textwidth]{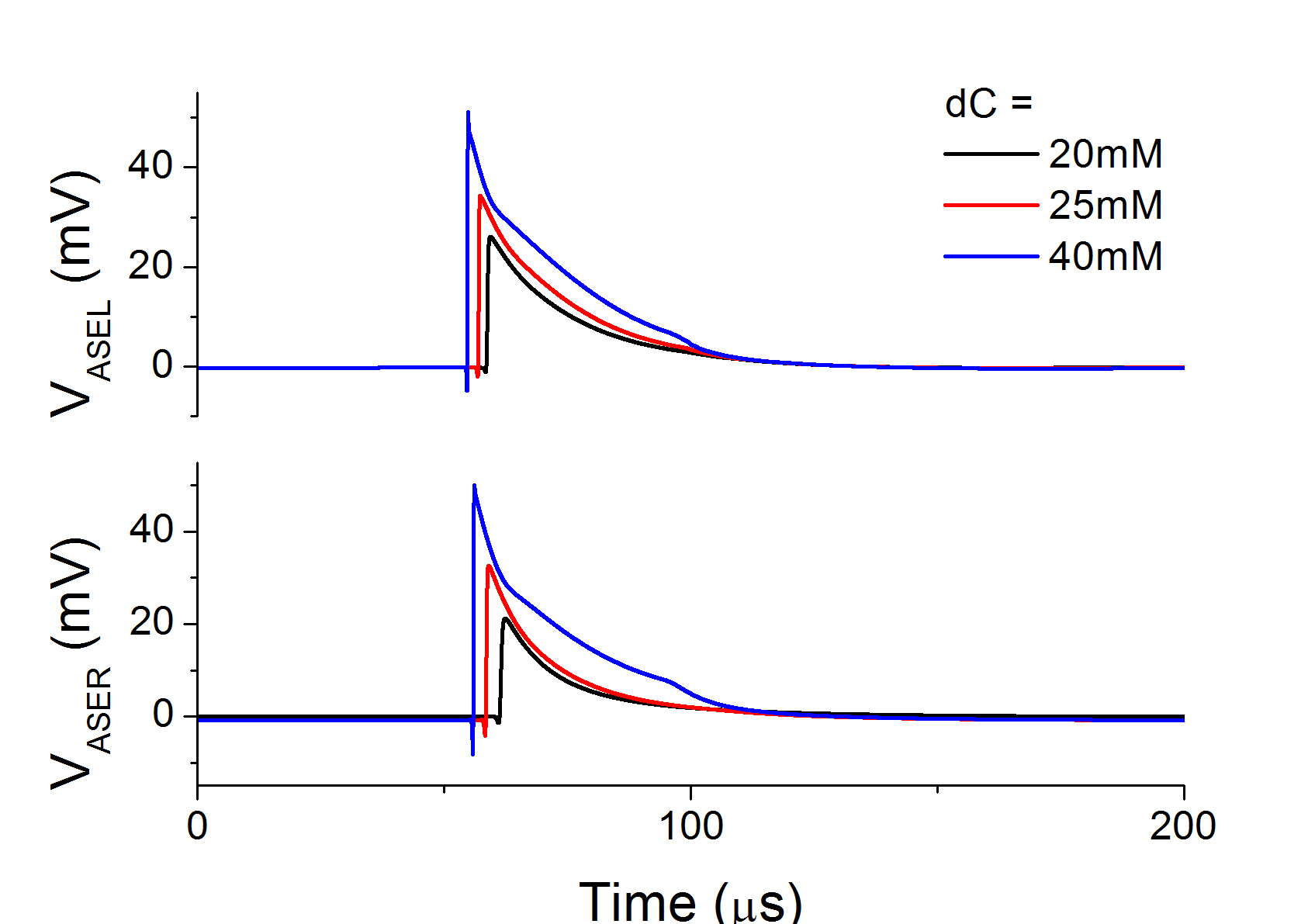}
    }
 \caption{
(a) Numerically calculated response of ASEL neuron to upstep in chemical concentration.
(b) Simulated Response of the VLSI circuit of ASEL neuron to upstep in chemical concentration.
\textit{Top:} Concentration profile of the input to the neuron \textit{Middle :} States of depolarising ion channels - $u^d, b^d$ and $i^d$ \textit{Bottom :} Output Potential of ASEL neuron
(c) Response of ASEL circuit to upsteps (\textit{Top}) and ASER circuit to downsteps (\textit{Bottom}) for different magnitudes from baseline concentration of $40\,$mM. Step-size is indicated by dC.
}
  \end{figure}

$\bullet$ \textit{\textbf{Membrane Potential Computation:}}\\
As described in Equation \ref{eq:cond}, the conductance dependent term $k^d$ for the depolarizing ion channels is $k_m \times (b^d)^2$. Hence computing $k^d$ is simply a matter of computing the square of a scaled version of the output current of the $b^d Block$, equivalently the current flowing through M34. This is done by the translinear circuit - $Squaring Block$ of Figure \ref{fig:c2}. To calculate the membrane potential of the ASEL neuron, Equation \ref{eq:pot} could be simplified by ignoring the terms corresponding the hyperpolarising ion channels since they are not present in ASEL and by setting the resting membrane potential to 0, as done for the numerically simulated neuron.
\begin{equation} \label{eq:redpot}
\tau_m \dot{V} =  k^d V_d  - V (k^d+ 1) 
\end{equation}
Comparing Equation \ref{eq:redpot} to Equation \ref{eq:de}, representing membrane potential as $I_{out}$, the translinear relationship required to implement this differential equation would be 
\begin{equation}
I_{out} \times (C \dot{V_x} + (k^d + 1) I_r) = (k^d V_d) \times I_r \\ 
\end{equation}
This gives $I_{in} = k^dV_d$ and $I_r = CV_t / k \tau_m$, where C and $I_r$ are suitably selected. $Output Block$ of Figure \ref{fig:c2} shows the circuit to generate the membrane potential of ASEL.
Figures \ref{fig:nr} and \ref{fig:cr} show the responses of the numerically simulated and circuit implementation of the ASEL neuron respectively. Though we have described only the circuit for the ASEL neuron here, a similar circuit for the ASER neuron could be obtained through minor modifications. Figure \ref{fig:steps} shows the response of designed circuits for ASEL and ASER neurons to upsteps and downsteps in concentration respectively. Our circuit, designed using $180\,$nm technology node operates at a supply voltage of $0.8\,$V and contains a total of $63$ transistors, $5$ capacitors ($285\,$fF) and consumes $\approx185\,$nW power. 
The developed neuron circuit could be easily converted to a spiking neuron via simple thresholding circuitry \cite{indiveri2011}.  The developed neuron pair could be integrated with existing circuits for LEIF neurons to develop a complete circuit for contour tracking and navigation.
\section{Conclusions}

In this paper, we have developed a spiking neural network inspired by chemotaxis in \textit{C. elegans}. Our network receives it\rq{}s input from a single concentration sensor and with merely $7$ spiking neurons in the network, our \lq\lq{}worm\rq\rq{} is able to explore, detect and track a desired concentration set-point. We have developed a pair of spiking neurons inspired by the ASE neurons in the \textit{C. elegans} to act as gradient detectors. Simulations show that our worm is able to detect the set-point with $\approx 4$ times higher probability than the optimal memoryless L\'{e}vy foraging model. Once the worm reaches the desired set-point, it tracks it with minimal deviation from the set-point. The developed network model is extremely versatile and changing the set-point or the temporal or spatial ranges over which the worm navigates is straightforward.

The developed network can be also modified to track other environmental variables such as temperature, radiation, etc by choosing appropriate sensors. The robustness of our model to noise strengthens its practical applicability. Our model could also be extended to three-dimensions since the worm only uses temporal gradients to make decisions and is oblivious to whether it\rq{}s motion is in two or three dimensions. The only modification required would be in determining the new position vector for the worm for which we would need to generate the turn angles appropriately to ensure three-dimensional motion. Our network responds in time scales similar to \textit{C. elegans} and all neurons spike at frequencies in the range of 1-10 Hz which would make for extremely energy efficient computation. We have also designed an analog sub-threshold CMOS circuit for capturing the graded potential dynamics of the ASE neurons accurately. The average power consumed by this circuit is $185\,$nW and it could be converted to a spiking circuit with already existing thresholding circuits. We are currently designing an integrated circuit with the other LEIF neurons to create a complete circuit for contour tracking and navigation. 
The developed algorithm and hardware could find applicability as a potentially power efficient and fault tolerant alternative to conventional navigation algorithms for robotic control.

\subsubsection*{References}
{\def\section*#1{}


\begin{thebibliography}{18}
\providecommand{\natexlab}[1]{#1}
\expandafter\ifx\csname urlstyle\endcsname\relax
  \providecommand{\doi}[1]{doi:\discretionary{}{}{}#1}\else
  \providecommand{\doi}{doi:\discretionary{}{}{}\begingroup
  \urlstyle{rm}\Url}\fi
\providecommand{\selectlanguage}[1]{\relax}

\small

\bibitem{Maass19971659}
Maass, W. (1997), Networks of spiking neurons: The third generation of neural
  network models, \emph{Neural Networks}, 10, 9, 1659--1671,
  \doi{http://dx.doi.org/10.1016/S0893-6080(97)00011-7}

\bibitem{Brenner01051974}
Brenner, S. (1974), The genetics of {C}aenorhabditis elegans, \emph{Genetics},
  77, 1, 71--94

\bibitem{White12111986}
White, J.~G., Southgate, E., Thomson, J.~N., and Brenner, S. (1986), The
  structure of the nervous system of the nematode caenorhabditis elegans,
  \emph{Philosophical Transactions of the Royal Society of London. B,
  Biological Sciences}, 314, 1165, 1--340, \doi{10.1098/rstb.1986.0056}

\bibitem{Ward01031973}
Ward, S. (1973), Chemotaxis by the nematode {C}aenorhabditis elegans:
  Identification of attractants and analysis of the response by use of mutants,
  \emph{Proceedings of the National Academy of Sciences}, 70, 3, 817--821,
  \doi{10.1073/pnas.70.3.817}

\bibitem{Dusenbery01051973}
Dusenbery, D.~B. (1973), Countercurrent separation: A new method for studying
  behavior of small aquatic organisms, \emph{Proceedings of the National
  Academy of Sciences}, 70, 5, 1349--1352

\bibitem{Bargmann1991729}
Bargmann, C.~I. and Horvitz., H. (1991), Chemosensory neurons with overlapping
  functions direct chemotaxis to multiple chemicals in {C}. elegans,
  \emph{Neuron}, 7, 5, 729--742,
  \doi{http://dx.doi.org/10.1016/0896-6273(91)90276-6}

\bibitem{Thiele23092009}
Thiele, T.~R., Faumont, S., and Lockery, S.~R. (2009), The neural network for
  chemotaxis to tastants in {C}aenorhabditis elegans is specialized for
  temporal differentiation, \emph{The Journal of Neuroscience}, 29, 38,
  11904--11911, \doi{10.1523/JNEUROSCI.0594-09.2009}

\bibitem{Miller30032005}
Miller, A.~C., Thiele, T.~R., Faumont, S., Moravec, M.~L., and Lockery, S.~R.
  (2005), Step-response analysis of chemotaxis in {C}aenorhabditis elegans,
  \emph{The Journal of Neuroscience}, 25, 13, 3369--3378,
  \doi{10.1523/JNEUROSCI.5133-04.2005}

\bibitem{PA}
Appleby, P. (2012), A model of chemotaxis and associative learning in {C}.
  elegans, \emph{Biological Cybernetics}, 106, 6-7, 373--387,
  \doi{10.1007/s00422-012-0504-8}

\bibitem{Suz}
Suzuki~Hiroshi, F. S. E. M. L. S.~R., Thiele Tod~R and R, S.~W. (2008),
  Functional asymmetry in {C}aenorhabditis elegans taste neurons and its
  computational role in chemotaxis, \emph{Nature}, 454, 114--117,
  \doi{10.1038/nature06927.}

\bibitem{kun}
Kunitomo, H., Sato, H., Iwata, R., Satoh, Y., Ohno, H., Yamada, K., et~al.
  (2013), Concentration memory-dependent synaptic plasticity of a taste circuit
  regulates salt concentration chemotaxis in {C}aenorhabditis elegans,
  \emph{Nature Communications}, 4, \doi{10.1007/s00422-012-0504-8}

\bibitem{Pierce-Shimomura01111999}
Pierce-Shimomura, J.~T., Morse, T.~M., and Lockery, S.~R. (1999), The
  fundamental role of pirouettes in {C}aenorhabditis elegans chemotaxis,
  \emph{The Journal of Neuroscience}, 19, 21, 9557--9569

\bibitem{Pierce-Shimomura15122005}
Pierce-Shimomura, J.~T., Dores, M., and Lockery, S.~R. (2005), Analysis of the
  effects of turning bias on chemotaxis in {C}. elegans, \emph{Journal of
  Experimental Biology}, 208, 24, 4727--4733, \doi{10.1242/jeb.01933}

\bibitem{Pappleby2013}
Appleby, P. (2013), The role of multiple chemotactic mechanisms in a model of
  chemotaxis in {C}. elegans: different mechanisms are specialised for
  different environments, \emph{Journal of Computational Neuroscience}, 1--16,
  \doi{10.1007/s10827-013-0474-4}

\bibitem{nolan:2015}
Nolan, J.~P. (2015), Stable Distributions - Models for Heavy Tailed Data
  (Birkhauser, Boston), in progress, Chapter 1 online at
  academic2.american.edu/$\sim$jpnolan

\bibitem{schaik2003}
Van~Schaik, A. and Jin, C. (2003), The tau-cell: a new method for the
  implementation of arbitrary differential equations, in Circuits and Systems,
  2003. ISCAS '03. Proceedings of the 2003 International Symposium on,
  volume~1, volume~1, I--569--I--572 vol.1, \doi{10.1109/ISCAS.2003.1205627}


\bibitem{indiveri2011}
Indiveri, G., Linares-Barranco, B., Hamilton, T.~J., Van~Schaik, A.,
  Etienne-Cummings, R., Delbruck, T., et~al. (2011), Neuromorphic silicon
  neuron circuits, \emph{Frontiers in neuroscience}, 5









\end{thebibliography}
\end{document}